\documentclass[10pt,twocolumn,letterpaper]{article}

\usepackage[pagenumbers]{cvpr} 

\definecolor{cvprblue}{rgb}{0.21,0.49,0.74}

\usepackage[breaklinks,colorlinks,allcolors=cvprblue]{hyperref} 

\usepackage{graphicx}
\usepackage{caption}
\usepackage{amsmath, amsfonts, amssymb}
\usepackage{colortbl}
\usepackage{pifont}
\usepackage{float}

\title{TED-VITON: Transformer-Empowered Diffusion Models for Virtual Try-On}

\author{
Zhenchen Wan$^1$ \quad Yanwu Xu \quad Zhaoqing Wang$^2$ \quad Feng Liu$^1$ \quad Tongliang Liu$^2$ \quad Mingming Gong$^{1,3}$\\
$^1$University of Melbourne, Melbourne, Australia \\
$^2$The University of Sydney, Sydney, Australia \\
$^3$Mohamed bin Zayed University of Artificial Intelligence, Abu Dhabi, UAE \\
{\tt\small zhenchenw@student.unimelb.edu.au, zwan6779@uni.sydney.edu.au, fengliu.ml@gmail.com} \\
{\tt\small tongliang.liu@sydney.edu.au, mingming.gong@unimelb.edu.au}
}

\begin{document}

\twocolumn[{%
\renewcommand\twocolumn[1][]{#1}%
\maketitle
\begin{center}
    \centering
    \includegraphics[width=\textwidth]{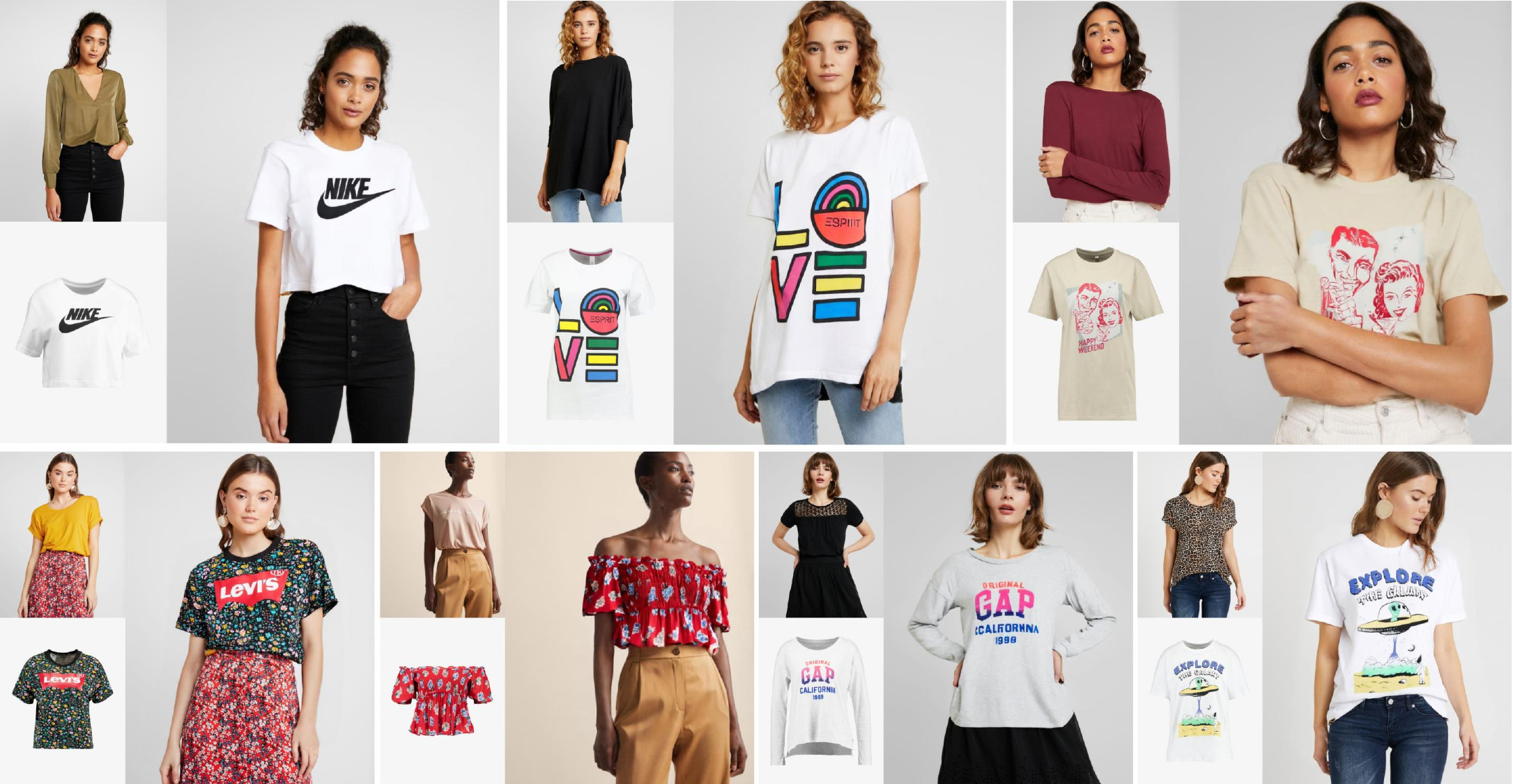}
    \captionsetup{type=figure, skip=-12pt}
    \captionof{figure}{We propose an implementation of Virtual Try-On using the Diffusion Transformer (DiT) architecture, which demonstrates state-of-the-art visual quality by preserving fine garment details and text clarity, even under challenging conditions involving complex human poses and diverse lighting environments.}
    \label{fig:cover_image}
\end{center}%
}]


\begin{abstract}

Recent advancements in Virtual Try-On (VTO) have demonstrated exceptional efficacy in generating realistic images and preserving garment details, largely attributed to the robust generative capabilities of text-to-image (T2I) diffusion backbones. However, the T2I models that underpin these methods have become outdated, thereby limiting the potential for further improvement in VTO. Additionally, current methods face notable challenges in accurately rendering text on garments without distortion and preserving fine-grained details, such as textures and material fidelity. The emergence of Diffusion Transformer (DiT) based T2I models has showcased impressive performance and offers a promising opportunity for advancing VTO. Directly applying existing VTO techniques to transformer-based T2I models is ineffective due to substantial architectural differences, which hinder their ability to fully leverage the models' advanced capabilities for improved text generation. To address these challenges and unlock the full potential of DiT-based T2I models for VTO, we propose TED-VITON, a novel framework that integrates a Garment Semantic (GS) Adapter for enhancing garment-specific features, a Text Preservation Loss to ensure accurate and distortion-free text rendering, and a constraint mechanism to generate prompts by optimizing Large Language Model (LLM). These innovations enable state-of-the-art (SOTA) performance in visual quality and text fidelity, establishing a new benchmark for VTO task. Project page: \url{https://zhenchenwan.github.io/TED-VITON/}

\end{abstract}

\section{Introduction}

Using input images of an individual and a selected garment, image-based VTO technology generates realistic images of the individual wearing the selected garment. By bypassing the necessity for physical fitting, VTO offers a transformative solution for applications in e-commerce, fashion cataloging, and the burgeoning metaverse. The primary challenges in VTO are threefold: (1) human body alignment, whereby the generated try-on image must accurately reflect the person’s body shape and pose; (2) garment fidelity, which preserves fine garment details, such as texture, color, and logo clarity, is essential to ensure authenticity; and (3) image quality, which pertains to the final output's resolution and the absence of artifacts.

While early VTO methods based on Generative Adversarial Networks (GANs) \cite{goodfellow_generative_2014} addressed these challenges to some extent \cite{bai_single_2022, choi_viton-hd_2021, dong_fw-gan_2019, han_viton_2018, minar_cp-vton_2020, wang_toward_2018, yang_towards_2020, yu_vtnfp_2019}, they often struggled with garment misalignment, visible artifacts, and limited generalizability. To address these limitations, diffusion models \cite{ho_denoising_2020} have emerged as a promising alternation in VTO research, leveraging a progressive noise-reversal process that enhances control over image generation and significantly improves texture and detail preservation. Recent UNet-based methods \cite{gou_taming_2023, morelli_ladi-vton_2023, ning_picture_2024, kim_stableviton_2023, li_unihuman_2023, wan_improving_2024, choi_improving_2024} utilize the generative strength of pretrained text-to-image (T2I) diffusion models \cite{rombach_high-resolution_2022, podell_sdxl_2023} to capture detailed garment semantics and enhance the realism of try-on images. These approaches achieve high image fidelity by encoding garment semantics through simple description \cite{morelli_ladi-vton_2023, choi_improving_2024} or using explicit warping networks \cite{gou_taming_2023, wan_improving_2024} to align garment structure with human poses. However, despite their advancements, these models still face challenges in preserving fine-grained garment details, such as logos, text, and intricate textures, and often struggle with accurately representing natural lighting and adapting to complex body poses.

To overcome these limitations, we explore the use of transformers in diffusion models, specifically building upon the DiT architecture~\cite{esser_scaling_2024}. Unlike UNet-based architectures, transformers offer enhanced scalability, long-range dependency modeling, and the ability to handle diverse visual contexts. However, directly migrating existing VTO approaches to the Transformer-based diffusion model does not guarantee performance improvements, as traditional UNet-based methods fail to fully exploit the potential of the transformer architecture. This observation aligns with our initial experiments, where a naive application of prior VTO techniques on DiT yielded suboptimal results.

To harness the capabilities of DiT, this paper proposes \textit{Transformer-Empowered Diffusion Models for Virtual Try-On} (TED-VITON), a framework designed to overcome key challenges in VTO by leveraging Transformer-based diffusion architectures. TED-VITON integrates several novel components to address limitations in garment detail preservation, model generalization, and text fidelity. Our contributions can be summarized as follows:

\begin{itemize}
    \item \textbf{Successful Migration of VTO to DiT-based Architecture:} We demonstrate the successful adaptation of Virtual Try-On technology to a DiT-based architecture. This paves the way for subsequent enhancements in the preservation of garment detail, semantic alignment, and visual fidelity.
    \item \textbf{Enhanced Garment Semantics with GS-Adapter:} Integrating the GS-Adapter, TED-VITON precisely aligns high-order semantic features from the image encoder with the DiT. This integration allows the model to more accurately capture occlusions, wrinkles, and material properties, maintaining realism across varied poses.
    \item \textbf{Text and Logo Clarity through Text Preservation Loss:} We introduce a Text Preservation Loss to address common challenges in text and logo fidelity. This loss function effectively enhances clarity and mitigates distortion, ensuring high-quality, distortion-free renderings of logos and text, critical for garments with complex designs.
    \item \textbf{Optimized Prompt Generation through Constraint Mechanism for LLMs:} To optimize DiT training, we introduce a constraint mechanism that tailors LLM prompts to garment-specific semantics. This mechanism improves training input quality, facilitating effective learning and generating outputs with superior visual fidelity.
\end{itemize}

\section{Related Works}

\noindent\textbf{Pose-Guided Person Image Synthesis (PPIS).} VTO technology originated with Pose-Guided Person Image Synthesis (PPIS). Initial PPIS approaches aimed to generate person images conditioned on specific body poses, laying the groundwork for generating visually convincing images of people in various postures. Pioneering works in this domain \cite{ma_pose_2017, liu_liquid_2019, zhu_progressive_2019, zhou_cross_2022, fruhstuck_insetgan_2022, albahar_pose_2021, li_pona_2020, men_controllable_2020} concentrated on aligning human poses with target clothing images, addressing key challenges in pose transfer and adapting to individual body shapes.

\noindent\textbf{GAN-based VTO.} Following PPIS advancements, VTO progressed to the application of Generative Adversarial Networks (GANs) for 2D VTO. GAN-based VTO approaches \cite{liu_liquid_2019, fruhstuck_insetgan_2022, kips_ca-gan_2020, dong_fw-gan_2019, honda_viton-gan_2019, raffiee_garmentgan_2021, pecenakova_fitgan_2022, albahar_pose_2021, men_controllable_2020, xie_gp-vton_2023, lee_high-resolution_2022, shim_towards_2024} typically involve two stages: deforming the garment to match the target person’s body shape, followed by fusing this deformed garment with the person’s image. Methods for improving garment deformation include using dense flow maps to create a seamless fit, while normalization and distillation techniques help to minimize misalignment. However, GAN-based VTO models face generalization limitations, especially in complex backgrounds and varied poses, limiting their applicability in dynamic real-world environments.

\noindent\textbf{Diffusion-based VTO.} Diffusion models have opened new avenues in VTO, enabling enhanced fidelity and detail preservation. Recent diffusion-based VTO methods \cite{cui_street_2024, bhatnagar_multi-garment_2019, zhu_tryondiffusion_2023, ning_picture_2024, li_unihuman_2023, morelli_ladi-vton_2023, gou_taming_2023, kim_stableviton_2023, wan_improving_2024, choi_improving_2024} extend beyond standard Stable Diffusion (SD), often employing customized architectures to boost performance. For instance, StableVITON \cite{kim_stableviton_2023} builds on SD1.4 \cite{rombach_high-resolution_2022} and incorporates ControlNet \cite{zhang_adding_2023} to enhance control over garment and body alignment, while IDM-VTON \cite{choi_improving_2024} leverages SDXL \cite{podell_sdxl_2023} with IP-Adapter \cite{ye_ip-adapter_2023} to refine garment-body fit through additional image-based control signals. These approaches effectively address key limitations of GAN-based methods, particularly in garment fidelity and preservation of fine details, establishing diffusion-based models as suitable for complex VTO applications. However, preserving intricate elements like garment text, logos, and texture under diverse poses and lighting conditions remains a challenge. TED-VITON aims to bridge these gaps, advancing VTO with a DiT architecture that integrates the GS-Adapter for semantic alignment, a DINOv2 encoder for capturing fine-grained garment details, and a Text Preservation Loss that ensures clarity in logos and text.

\section{Methodology}

\begin{figure*}[ht]
    \centering
    \includegraphics[width=1\textwidth]{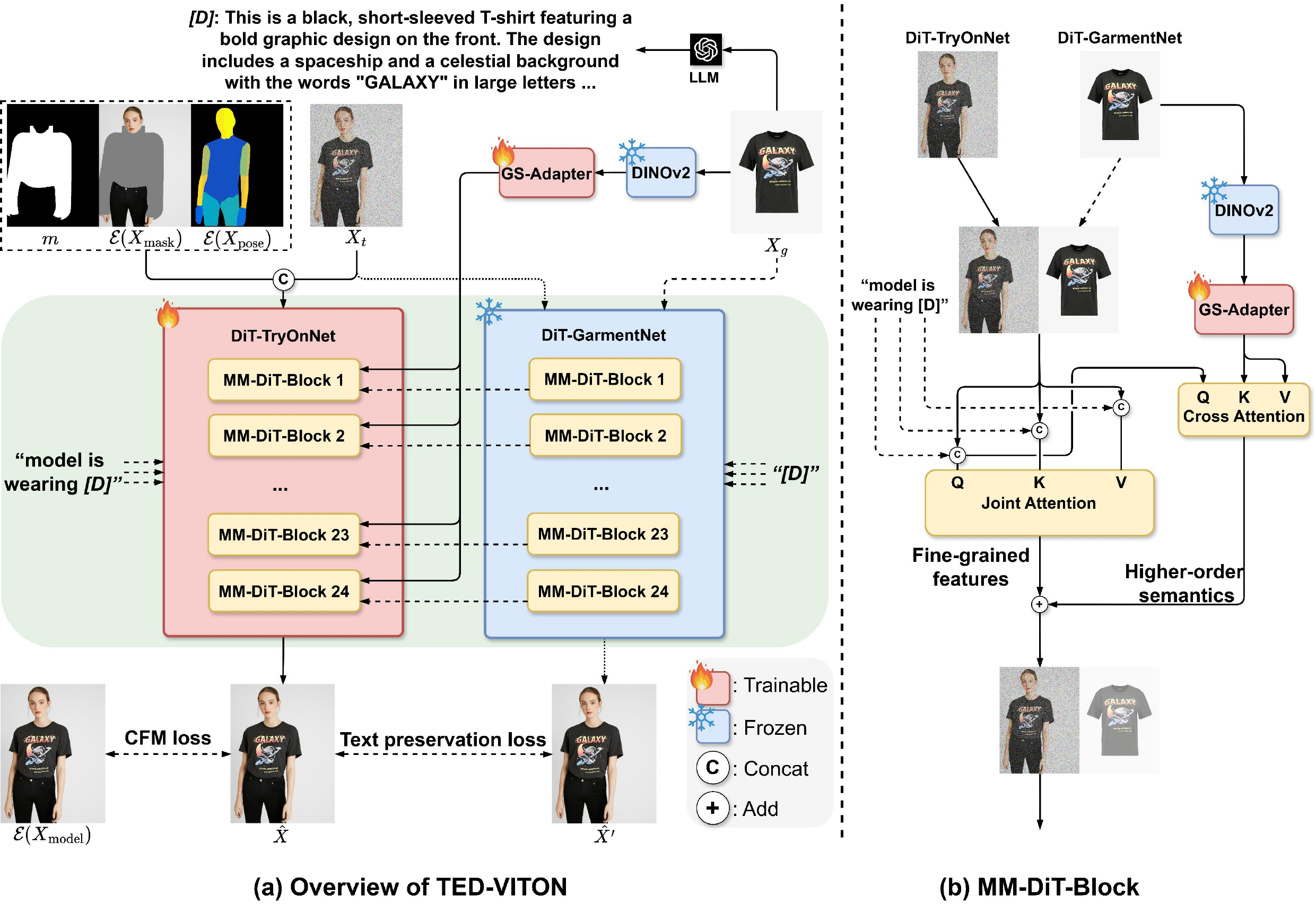}
    \caption{\textbf{Overview of TED-VITON:} We present the architecture of the proposed model along with details of its block modules. \textbf{(a)} Our model consists of 1) DiT-GarmentNet that encodes fine-grained features of \( X_g \), 2) GS-Adapter \cite{ye_ip-adapter_2023} that captures higher-order semantics of garment image \( X_g \), and 3) DiT-TryOnNet, the main Transformer for processing person images. The Transformer input is formed by concatenating the noised latents \( X_t \) with the segmentation mask \( m \), masked image \( \mathcal{E}(X_\text{model}) \), and Densepose \cite{guler_densepose_2018} \( \mathcal{E}(x_{\text{pose}}) \). Additionally, a detailed description of the garment (e.g., “[D]: The clothing item is a black T-shirt...”) is generated through an LLM and fed as input to both the DiT-GarmentNet and DiT-TryOnNet. The model aims to preserve garment-specific details through a text preservation loss, which ensures that key textual features are retained. \textbf{(b)} Intermediate features from DiT-TryOnNet and DiT-GarmentNet are concatenated. These are then refined through joint-attention and cross-attention layers, with the GS-Adapter further contributing to the refinement process. In this architecture, the DiT-TryOnNet and GS-Adapter modules are fine-tuned, while other components remain frozen.}
    \label{fig:TED-VITON}
\end{figure*}

\subsection{Background on Controlling Diffusion Models}

\textbf{Stable Diffusion (SD) 3 Model.} The SD 3 model \cite{esser_scaling_2024} represents a significant breakthrough as the first diffusion model to utilize a Transformer-based architecture. Building upon Latent Diffusion Models (LDMs) \cite{rombach_high-resolution_2022}, SD3 introduces the rectified flow approach \cite{liu_flow_2022}, which connects data points in the latent space via straight linear paths, replacing the traditional curved trajectories. This straight-line trajectory minimizes noise accumulation and allows for efficient, high-quality image synthesis.

In SD3, the input image \( x \) is encoded into a latent representation \( z_0 = E(x) \) by a pre-trained encoder \( E \). The rectified flow formulation defines a forward diffusion process with a variance schedule \( \beta_t \) expressed as follows:

\begin{equation}
q(z_t | z_0) = \mathcal{N}(z_t; \sqrt{\bar{\alpha}_t} z_0, (1 - \bar{\alpha}_t)I),
\end{equation}

\noindent where \( t \in \{1, \dots, T\} \) indicates diffusion steps, \( \alpha_t := 1 - \beta_t \), and \( \bar{\alpha}_t := \prod_{s=1}^{t} \alpha_s \). SD3 leverages the Conditional Flow Matching (CFM) loss to guide rectified flow during training:

\begin{equation}
\mathcal{L}_{\text{CFM}} = \mathbb{E}_{E(x), \epsilon \sim \mathcal{N}(0, 1), t} \left[ \left\| v_\theta(z, t) - u_t(z | \epsilon) \right\|^2 \right],
\label{CFM}
\end{equation}

\noindent where \( u_t(z | \epsilon) \) denotes the rectified vector field for direct, linear alignment. Unlike prior diffusion models, SD3 employs a Transformer backbone (Multimodal DiT) that facilitates bidirectional information flow between text and image tokens. This multimodal structure enhances text comprehension and visual quality, making SD3 highly suitable for text-guided image generation with improved fidelity and detail retention.

\noindent\textbf{ControlNet for Conditional Image Generation.} ControlNet \cite{zhang_adding_2023} extends diffusion models by enabling conditional image generation with additional guidance inputs, such as edge maps, segmentation masks, or pose annotations. ControlNet operates by branching from the base model’s intermediate features \( F \). The conditional input \( C \) is processed with learnable weights \( W_c \), resulting in conditioned features \( F_{\text{ctrl}} = \text{ControlNet}(C; W_c) \). These conditioned features are merged back into the main pipeline as \( F_{\text{combined}} = F + \lambda F_{\text{ctrl}} \), where \( \lambda \) regulates the influence of the conditional input. During training, ControlNet minimizes a composite loss:

\begin{equation}
\mathcal{L}_{\text{control}} = \mathcal{L}_{\text{diff}} + \gamma \mathcal{L}_{\text{cond}},
\end{equation}

\noindent where \( \mathcal{L}_{\text{diff}} \) is the base diffusion model’s loss, \( \mathcal{L}_{\text{cond}} \) ensures alignment with the guidance input, and \( \gamma \) balances their contributions. This framework enables precise control over image generation, making ControlNet highly effective for tasks requiring fine-grained customization.

\subsection{TED-VITON}
Figure \ref{fig:TED-VITON} (a) illustrates the TED-VITON framework, comprising DiT-GarmentNet, the Garment Semantic (GS) Adapter and DiT-TryOnNet. The following section provides a comprehensive description of each module and the training procedure.

\noindent\textbf{DiT-GarmentNet.} DiT-GarmentNet is designed to extract fine-grained garment features, including textures, patterns, fabric structures, logos, and other subtle design elements essential for realistic VTO results. By preserving the garment's true visual characteristics, this module ensures high fidelity, particularly in applications requiring precise appearance rendering.

DiT-GarmentNet processes the latent representation of the garment image \( \mathcal{E}(X_g) \), extracted via a pre-trained VAE encoder \(\mathcal{E}\), along with the conditioned text prompt \( \tau_\theta(D) \) generated by a multi-modal text encoders. These representations flow through multiple transformer layers, refining and retaining intricate garment details. The transformer architecture, inspired by \citet{esser_scaling_2024}, captures long-range dependencies, ensuring consistent textures, accurate logo placement.

DiT-GarmentNet processes the garment image \( X_g \) alongside the conditioned text prompt \( D \), is defined as:
\begin{equation}
    F_{\text{garment}}^i = \text{DiT-GarmentNet}^i(\mathcal{E}(X_g), \tau_\theta(D)),
\end{equation}
\noindent where \( F_{\text{garment}}^i \) denotes the fine-grained features extracted from the \( i \)-th transformer layer of DiT-GarmentNet.

In this way, DiT-GarmentNet ensures high visual fidelity by combining garment-specific details with broader model context, enabling the VTO system to accurately render complex designs on various body shapes and poses.

\noindent\textbf{Garment Semantic Adapter (GS-Adapter).} The GS-Adapter \cite{ye_ip-adapter_2023} is a key module that enhances generalization, making the model less sensitive to variations in body poses, garment deformations, and conditions like lighting or camera angles. By focusing on low-frequency features, it captures essential garment attributes, enabling consistent performance across diverse scenarios.

Unlike DiT-GarmentNet, which extracts high-frequency details like textures and logos, the GS-Adapter uses the DINOv2 encoder \cite{oquab2023dinov2} to distill  semantic garment information, including structure, style, and material. These high-order semantics, \( H_{\text{semantic}} \), encapsulate broader contextual attributes while maintaining adaptability.

The GS-Adapter employs a decoupled cross-attention mechanism to independently process joint and image embeddings. Let \( \mathbf{Q} \in \mathbb{R}^{N \times d} \) represent the query matrix, and \( \mathbf{K}_j, \mathbf{V}_j \) and \( \mathbf{K}_i, \mathbf{V}_i \) denote key-value pairs for joint and image embeddings, respectively. The combined output is:

\begin{equation}
    \mathbf{Z}_{\text{new}} = \text{Attention}(\mathbf{Q}, \mathbf{K}_j, \mathbf{V}_j) + \lambda \cdot \text{Attention}(\mathbf{Q}, \mathbf{K}_i, \mathbf{V}_i),
    \label{eq:GS-Adapter}
\end{equation}

\noindent where \( \lambda \) balances image and joint feature contributions. This design allows the GS-Adapter to generalize effectively across diverse poses, complex garments, and varying environmental conditions, enhancing model robustness and ensuring realistic outputs.

\noindent\textbf{DiT-TryOnNet.} DiT-TryOnNet builds upon the DiT architecture, leveraging its powerful Transformer-based diffusion capabilities within the latent space of a pre-trained VAE. By integrating DiT, our model benefits from the scalability and long-range dependency modeling of Transformers, enabling precise alignment and realistic rendering in virtual try-on scenarios. For DiT-TryOnNet, we construct a combined input \( \zeta = [\mathcal{E}(X_{\text{model}}); m; \mathcal{E}(X_{\text{mask}}); \mathcal{E}(X_{\text{pose}})] \) to provide a comprehensive context. This input consists of: the person’s latent image representation \( \mathcal{E}(X_{\text{model}}) \) as the primary structural guide; a dynamically resized mask \( m \) to isolate the garment area and focus the model’s attention; the masked person’s image \( X_{\text{mask}} = (1 - m) \odot X_{\text{model}} \) for garment reconstruction; and the DensePose embedding \( \mathcal{E}(X_{\text{pose}}) \) to align with the person’s pose.

Within the \textbf{MM-DiT-Block} (Fig.~\ref{fig:TED-VITON}(b)), fine-grained garment details \(F_{\text{garment}}^i\) extracted from the \(i\)-th transformer layer of DiT-GarmentNet, merge with the feature representation \(F_{\text{tryon}}^i\) from the corresponding \(i\)-th layer DiT-TryOnNet to form \(F_{\text{image}}^i\), which serves as the primary input for attention processing. Descriptive text embeddings \(\tau_\theta(D)\), generated by multimodal text encoders, are concatenated with \(F_{\text{image}}^i\) within the query, key, and value components of the joint attention mechanism (i.e., \(Q_{\text{joint}} = \text{Concat}(Q_{\text{image}}^i, Q_{\tau_\theta(D)})\), \(K_{\text{joint}} = \text{Concat}(K_{\text{image}}^i, K_{\tau_\theta(D)})\), \(V_{\text{joint}} = \text{Concat}(V_{\text{image}}^i, V_{\tau_\theta(D)})\)). This results in a hidden state \(H_{\text{joint}}^i\) that unifies visual and textual modalities. Subsequently, this hidden state is further enriched by incorporating high-order semantic features, \( H_{\text{semantic}} \) provided by the GS-Adapter, as described in Eq. \ref{eq:GS-Adapter}.

To produce the final VTO output \( \hat{X} \), DiT-TryOnNet leverages the combined input \( \zeta \) and garment description embedding \( \tau_\theta(D) \):

\begin{equation}
    \hat{X} = \text{DiT-TryOnNet}(\zeta, \tau_\theta(D)).
    \label{eq:tryonnet_output}
\end{equation}

\noindent\textbf{Prior Preservation for Text Generation.} To retain the model’s ability to generate accurate and clear text, such as logos and labels, we introduce a prior preservation mechanism inspired by DreamBooth \cite{ruiz_dreambooth_2023}. This mechanism incorporates a text preservation loss to ensure text clarity and fidelity, preventing the model from losing this capability while fine-tuning for VTO tasks. As the final component of our framework, prior preservation complements the GS-Adapter and DiT-TryOnNet. Together, they form a comprehensive training objective, achieving a balance between high-fidelity garment rendering and robust text generation for realistic VTO outputs.

As shown in Fig.~\ref{fig:TED-VITON}(a), the total loss function combines two main components: (1) the CFM loss \( \mathcal{L}_{\text{CFM}} \) defined in Eq.~\ref{CFM}, which ensures high-quality VTO outputs by aligning generated images with the desired garment and pose, and (2) the text preservation loss \( \mathcal{L}_{\text{pres}} \), which maintains clarity in text details. The CFM loss guides the model in generating the VTO result \( \hat{X} \) leveraging DiT-GarmentNet for detail retention and DiT-TryOnNet for fit adjustments based on pose and body type. The text preservation loss \( \mathcal{L}_{\text{pres}} \) is computed as \( \mathcal{L}_{\text{pres}} = \text{MSE}(\hat{X}, \hat{X}') \), where \( \hat{X}' \) is the baseline latent representation from the original model, helping to retain text fidelity in the fine-tuned output. The final loss function is given by:

\begin{equation}
    \mathcal{L}_{\text{total}} = \mathcal{L}_{\text{CFM}} + \lambda_{\text{pres}} \cdot \mathcal{L}_{\text{pres}},
    \label{eq:preservation}
\end{equation}

\noindent where \( \lambda_{\text{pres}} \) controls the balance between VTO adaptation and text retention. This approach enables high-quality garment realism while preserving essential text rendering for realistic try-on images.

\noindent\textbf{GPT-4o Generated Garment Descriptions.} Our approach uses GPT-4o to generate detailed garment descriptions that capture both basic and nuanced features. These descriptions provide rich semantic context, enhancing the model's ability to faithfully represent garment details. For \textbf{DiT-GarmentNet}, descriptions help preserve intricate details like texture and logos. Meanwhile, for \textbf{DiT-TryOnNet}, the text prompt is tailored to emphasize how the garment appears when worn, focusing on fit and interaction with the body. This adjustment improves realism in the generated images. This dual-conditioning approach enables more accurate garment representation, as shown in Fig.~\ref{fig:TED-VITON}(a).

\begin{figure*}[ht]
    \centering
    \includegraphics[width=\textwidth]{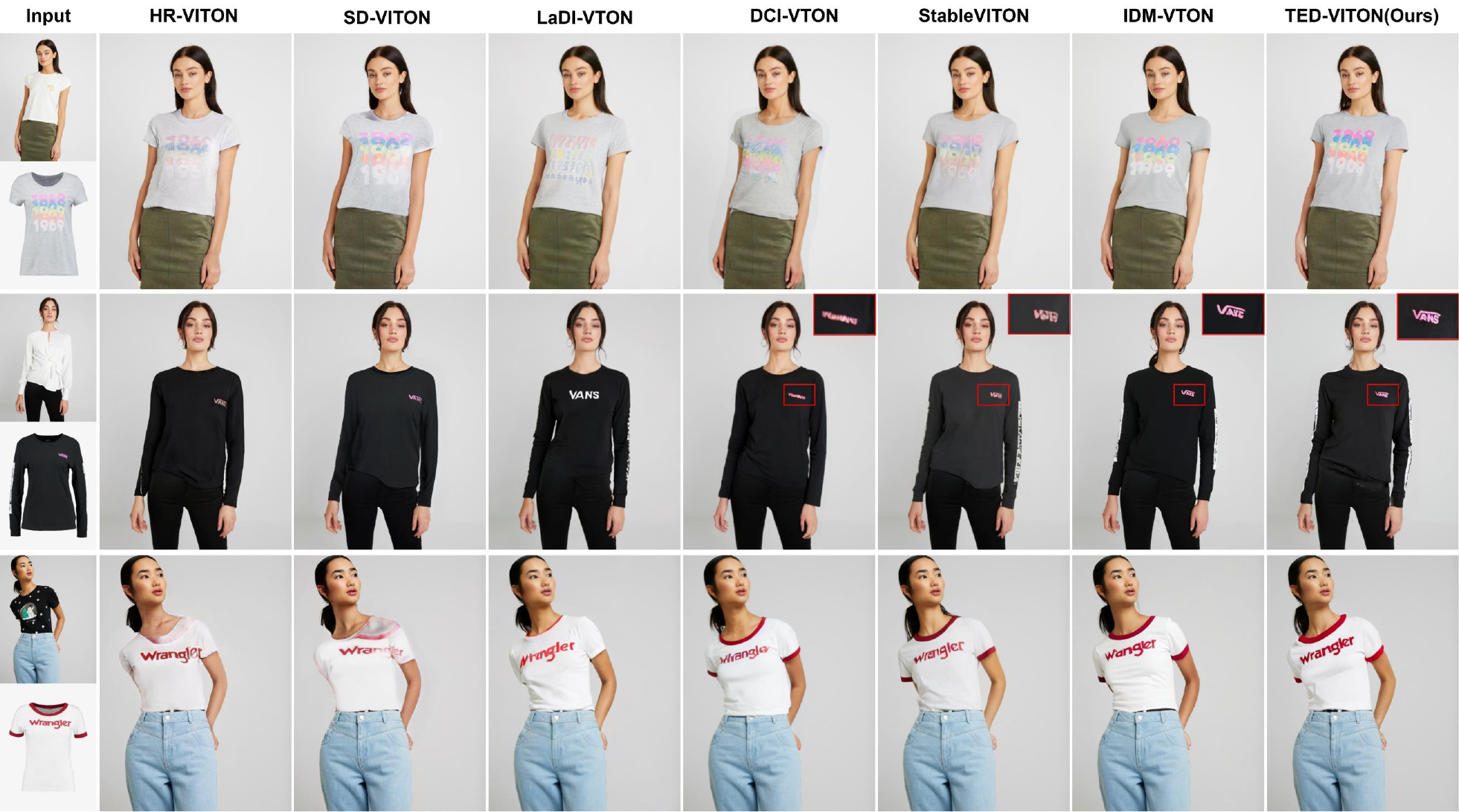}
        \caption{Qualitative comparison with baseline methods. This figure demonstrates the superior performance of our model compared to various SOTA approaches. Zooming in reveals finer details.}
    \label{fig:Qualitative_comparison}
\end{figure*}

\begin{table*}[ht]
\centering
\vspace{-0.3em}
\resizebox{1\textwidth}{!}{
\begin{tabular}{lccccclccccc}
\toprule
\textbf{Dataset} & \multicolumn{5}{c}{\textbf{VITON-HD}} &  & \multicolumn{5}{c}{\textbf{DressCode Upper-body}} \\ \cline{2-6} \cline{8-12} 
\textbf{Method} & LPIPS↓ & SSIM↑ & \multicolumn{1}{l}{CLIP-I↑} & FID↓ UN & KID↓ UN &  & LPIPS↓ & SSIM↑ & \multicolumn{1}{l}{CLIP-I↑} & FID↓ UN & KID↓ UN \\ \hline
\rowcolor[gray]{0.9} \multicolumn{12}{c}{\textbf{GAN-based methods}} \\
\hline
\textbf{HR-VITON \cite{lee_high-resolution_2022}} & 0.115 & 0.877 & 0.800 & 12.238 & 3.757 &  & 0.118 & 0.910 & 0.749 & 29.383 & 3.104 \\
\textbf{SD-VITON \cite{shim_towards_2024}} & 0.104 & \textbf{0.896} & 0.831 & 9.857 & 1.450 &  & - & - & - & - & - \\ \hline
\rowcolor[gray]{0.9} \multicolumn{12}{c}{\textbf{Diffusion-based methods}} \\
\hline
\textbf{LaDI-VTON \cite{morelli_ladi-vton_2023}} & 0.166 & 0.873 & 0.819 & 9.386 & 1.590 &  & 0.157 & 0.905 & 0.789 & 22.689 & 2.580 \\
\textbf{DCI-VTON \cite{gou_taming_2023}} & 0.197 & 0.863 & 0.823 & 9.775 & 1.762 &  & 0.171 & 0.893 & 0.756 & 24.184 & 2.379 \\
\textbf{StableVITON \cite{kim_stableviton_2023}} & 0.142 & 0.875 & 0.838 & 9.371 & 1.990 &  & 0.113 & 0.910 & 0.844 & 19.712 & 2.149 \\
\textbf{IDM–VTON \cite{choi_improving_2024}} & \underline{0.102} & 0.868 & \underline{0.875} & \underline{9.156} & \underline{1.242} &  & \underline{0.065} & \underline{0.920} & \underline{0.870} & \underline{11.852} & \textbf{1.181} \\
\textbf{TED-VITON (Ours)} & \textbf{0.095} & \underline{0.881} & \textbf{0.878} & \textbf{8.848} & \textbf{0.858} &  & \textbf{0.050} & \textbf{0.934} & \textbf{0.875} & \textbf{11.451} & \underline{1.393} \\
\bottomrule
\end{tabular}
}
\caption{Quantitative comparison on training models on VITON-HD and evaluate them on both VITON-HD and DressCode upper-body datasets. \textbf{Bold} and \underline{underline} denote the best and the second best result, respectively. ``UN'' indicated the unpaired setting. KID score is multiplied by 100.}
\label{tab:result}
\vspace{-0.5em}
\end{table*}

\section{Experiment}

To thoroughly evaluate TED-VITON, we conduct a comprehensive study that includes quantitative and qualitative analyses, ablation studies to assess the contributions of individual components, and a user study to gauge human preferences. For the quantitative analysis, we measure standard metrics to evaluate the generated images' alignment with ground truth and overall visual quality. In the qualitative analysis, we compare TED-VITON's outputs with those of baseline models to examine its ability to reproduce fine garment features, such as textures, logos, and material details. We also perform ablation studies by systematically removing key components to assess their impact on performance and image quality. Since human preference is a critical measure of success in generative tasks, we conduct a user study to gather feedback on the perceived realism and aesthetic appeal of the fitting images. The results highlight TED-VITON's ability to produce visually compelling and realistic outputs that surpass existing methods.

\subsection{Experiment Setup}

\noindent\textbf{Baselines.} We evaluate our method against both GAN-based and diffusion-based VTO approaches. The GAN-based baselines include HR-VITON \cite{lee_high-resolution_2022} and SD-VTON \cite{shim_towards_2024}. Both methods employ a separate warping module to fit the garment onto the target person, followed by GAN-based generation with the fitted garment as input. Among the diffusion-based methods, we compare with LaDI-VTON \cite{morelli_ladi-vton_2023}, DCI-VTON \cite{gou_taming_2023}, StableVITON \cite{kim_stableviton_2023}, and IDM-VTON \cite{choi_improving_2024}. All of these models leverage pretrained SD models, though with different conditioning techniques. LaDI-VTON and DCI-VTON incorporate distinct warping modules for garment conditioning, while StableVITON directly uses the SD1.4 encoder for conditioning. IDM-VTON, by contrast, utilizes the SDXL inpainting model checkpoints from official repositories. Our approach similarly builds on SD3 original checkpoints from official sources. For a fair comparison, we generate images at a resolution of \(1024 \times 768\) when available; otherwise, we generate images at \(512 \times 384\) and upscale them to \(1024 \times 768\) using interpolation or super-resolution techniques \cite{wang_real-esrgan_2021}, reporting the highest-quality results achieved.

\begin{figure*}[t]
    \centering
    \includegraphics[width=0.93\textwidth]{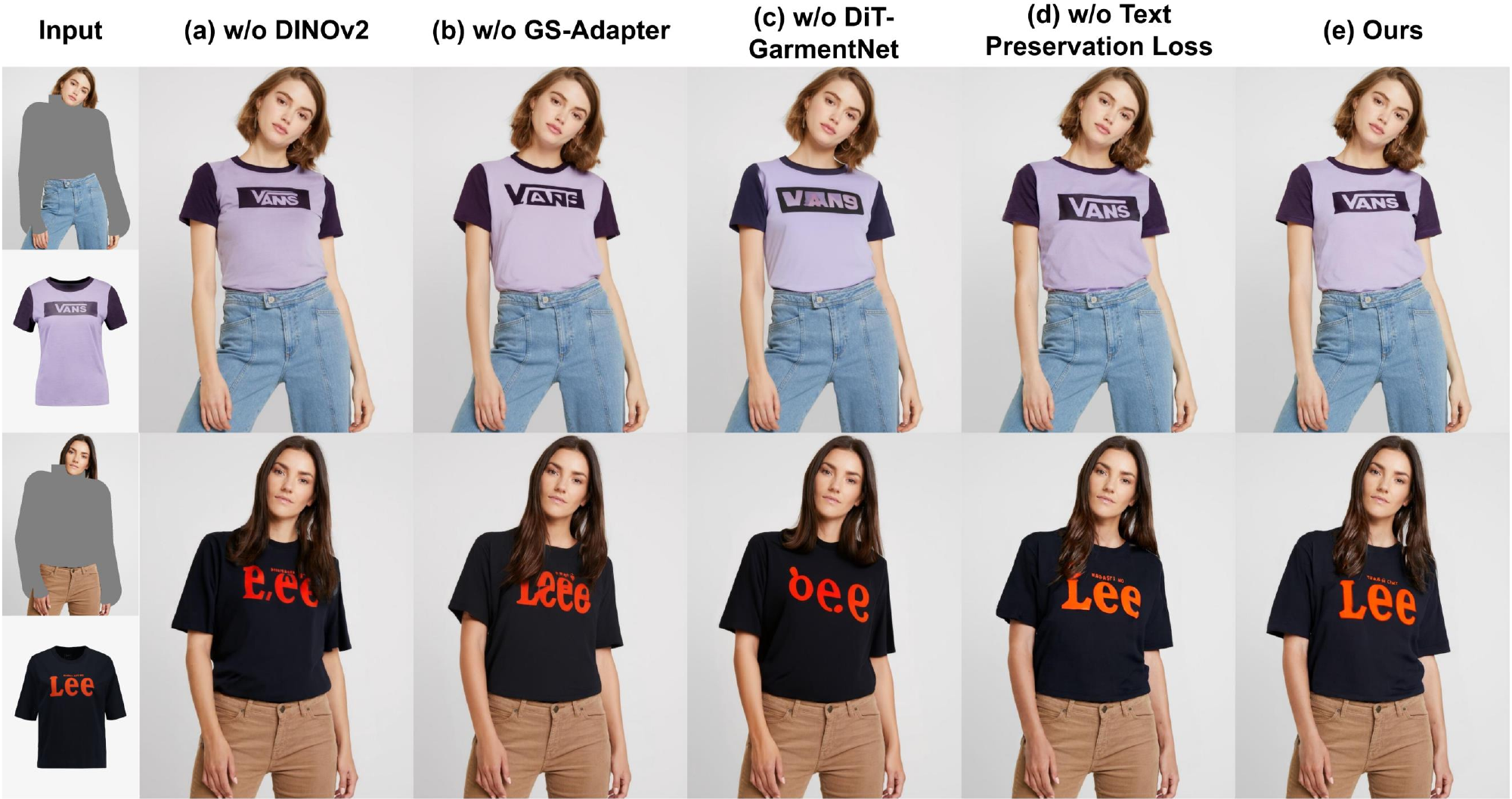}
    \caption{An ablation study on the key components of TED-VITON.}
    \label{fig:ablation_study}
    \vspace{-1em}
\end{figure*}

\begin{table}[t]
    \centering
    \resizebox{1\linewidth}{!}{
        \begin{tabular}{lcccccc}
            \toprule
            \textbf{Component} & \textbf{LPIPS↓} & \textbf{SSIM↑} & \textbf{CLIP-I↑} & \textbf{FID↓ UN} & \textbf{KID↓ UN} \\ 
            \hline
            \textbf{w/o DINOv2} & 0.120 & 0.870 & 0.852 & 9.655 & 1.734 \\
            \textbf{w/o GS-Adapter} & 0.111 & 0.842 & 0.829 & 9.674 & 1.680 \\
            \textbf{w/o DiT-GarmentNet} & 0.113 & 0.850 & 0.817 & 9.931 & 1.693 \\
            \textbf{w/o Text Preservation Loss} & 0.098 & 0.877 & 0.864 & 9.438 & 1.487 \\
            \hline
            \textbf{Full Model} & \textbf{0.095} & { \textbf{0.881}} & { \textbf{0.878}} & \textbf{8.848} & \textbf{0.858} \\
            \bottomrule
        \end{tabular}
    }
    \caption{Quantitative comparison of models trained on the VITON-HD dataset with and without each component. ``UN'' indicated the unpaired setting.  KID score is multiplied by 100.}
    \label{tab:ablation_study_combined}
    \vspace{-1em}
\end{table}

\noindent\textbf{Evaluation datasets.} We evaluate the effectiveness of TED-VITON on two widely-used VTO datasets, VITON-HD \cite{choi_viton-hd_2021} and DressCode \cite{morelli_dress_2022}. The VITON-HD dataset consists of 13,679 pairs of frontal-view images of women and corresponding upper garments. Following the standard dataset practices of previous works \cite{morelli_ladi-vton_2023, gou_taming_2023, kim_stableviton_2023, choi_improving_2024, wan_improving_2024}, we divide VITON-HD into a training set of 11,647 pairs and a test set of 2,032 pairs. The DressCode dataset contains 15,366 image pairs focused specifically on upper-body garments. Consistent with the original dataset splits, we use 1,800 upper-body image pairs from DressCode as the test set. All experiments on both VITON-HD and DressCode are conducted at a resolution of \(1024 \times 768\).

\noindent\textbf{Evaluation metrics.} We evaluate TED-VITON in both paired and unpaired settings, following established practices in VTO literature. In the paired setting, the input garment matches the one originally shown in the person image. To assess performance, we use three key metrics: Structural Similarity Index (SSIM) \cite{wang_image_2004}, Learned Perceptual Image Patch Similarity (LPIPS) \cite{zhang_unreasonable_2018} and the CLIP image similarity score (CLIP-I) \cite{hessel_clipscore_2021} to measure similarity between the generated image and the ground truth. Additionally, in the unpaired setting, where the garment in the person image is replaced with a different one and no ground truth is available, we assess TED-VITON’s performance in terms of image quality and realism using Fréchet Inception Distance (FID) \cite{heusel_gans_2017} and Kernel Inception Distance (KID) \cite{kim_u-gat-it_2019} scores.

\subsection{Qualitative Results}

Fig.~\ref{fig:Qualitative_comparison} provides a qualitative comparison of VTO models alongside the input person image and selected garments. TED-VITON stands out as the only model capable of accurately reproducing text details on garments, such as the large ``1969'' and ``Wrangler'' logos, as well as finer text like ``Vans''. In terms of color and texture fidelity, TED-VITON precisely aligns the four colors in ``1969'' across the text rows, maintaining the original garment’s design. Unlike other models, which often exhibit text distortion or color misalignment, TED-VITON preserves text clarity and color accuracy. This is achieved through the integration of a Text Preservation Loss and enhanced prompt conditioning, which together ensure that fine-grained text and color details are retained in the generated VTO images.

\subsection{Quantitative Results}

\noindent\textbf{VITON-HD.} We evaluate TED-VITON on the VITON-HD dataset and compare it with SOTA VTO methods, including GAN-based approaches (HR-VITON \cite{lee_high-resolution_2022} and SD-VITON \cite{shim_towards_2024}) and diffusion-based methods (LaDI-VTON \cite{morelli_ladi-vton_2023}, DCI-VTON \cite{gou_taming_2023}, StableVITON \cite{kim_stableviton_2023}, and IDM-VTON \cite{choi_improving_2024}). Table \ref{tab:result} presents the quantitative results, where TED-VITON achieves top scores in LPIPS, CLIP-I, FID, and KID, indicating superior perceptual quality and realism. It ranks second in SSIM, highlighting its strong structural similarity preservation and alignment with perceptual semantics.

\noindent\textbf{DressCode Upper-body.} To evaluate TED-VITON's generalization across diverse garment styles, we test it on the DressCode upper-body dataset. As shown in Table \ref{tab:result}, TED-VITON outperforms other models across most metrics, achieving top scores in LPIPS, SSIM, CLIP-I and FID scores, which indicate strong alignment with the perceptual features of the in-shop garment. TED-VITON outperforms diffusion-based models like IDM-VTON and StableVITON by consistently capturing finer patterns and more accurate garment textures. In contrast, GAN-based methods struggle with complex patterns, resulting in lower-quality outputs on this dataset.

\subsection{User Study}

\begin{figure}[t]
    \centering
    \includegraphics[width=1\linewidth]{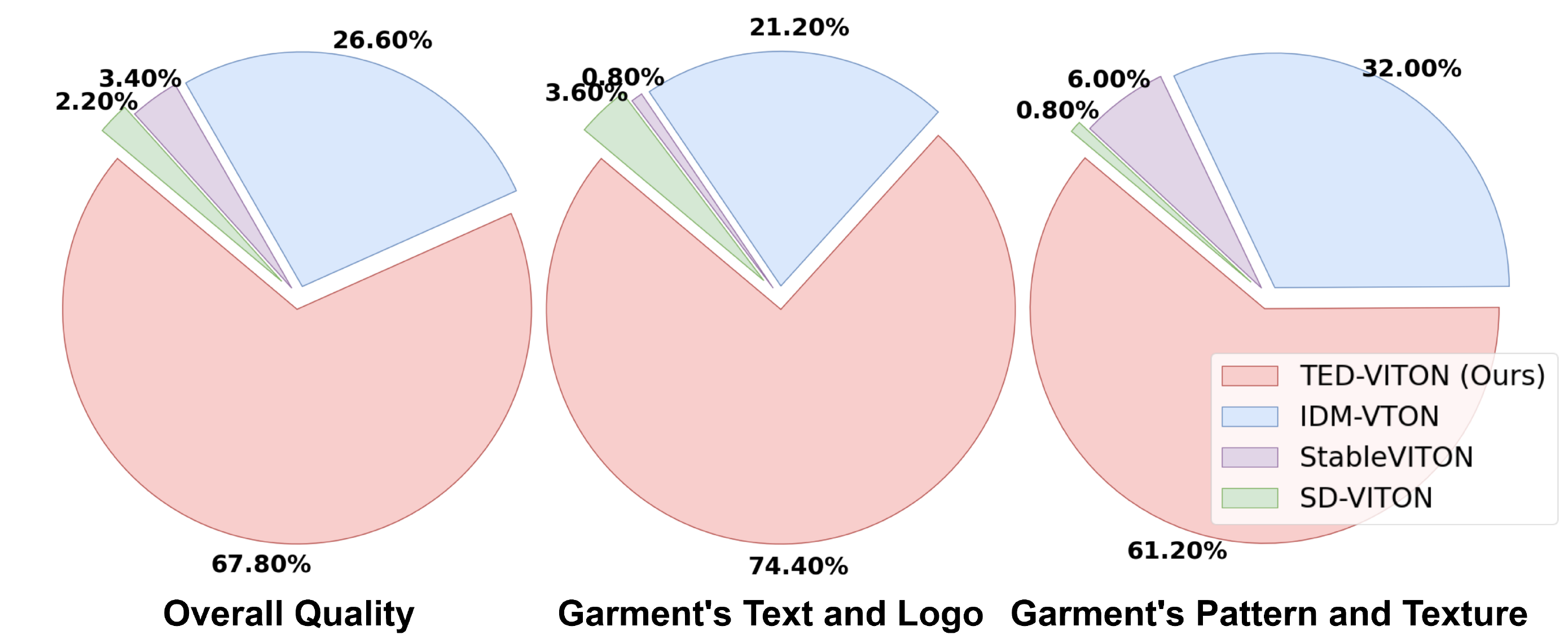}
    \caption{User study results based on 10 selected pairs from the VITON-HD \cite{choi_viton-hd_2021} dataset: 5 pairs assessed text and logo preferences, and the other 5 focused on pattern and texture preferences.}
    \label{fig:user_study}
\vspace{-0.4cm}
\end{figure}

To complement objective metrics, we conducted a user study to evaluate the visual appeal of our model. The study used 10 image pairs from the VITON-HD dataset, divided into two groups: 5 pairs focused on text and logo clarity, and 5 pairs evaluated pattern and texture fidelity. As shown in Fig.~\ref{fig:user_study}, with 50 valid responses, TED-VITON was the preferred model, demonstrating strong user preference for text clarity and pattern accuracy.

\subsection{Ablation Study}

\noindent\textbf{Effect of key components of TED-VITON.} To analyze the contributions of each key component in TED-VITON, we conduct an ablation study by systematically removing individual components, such as DINOv2, the GS-Adapter, DiT-GarmentNet, or Text Preservation Loss, and evaluate their impact on the results. In Fig.~\ref{fig:ablation_study}(a), replacing DINOv2 with a standard CLIP encoder results in blurry and distorted text. This highlights DINOv2’s role, in conjunction with the GS-Adapter, in enhancing text clarity and garment alignment by capturing fine semantic and garment details. Fig.~\ref{fig:ablation_study}(b) demonstrates the impact of removing the GS-Adapter, leading to misaligned garment features and reinforcing its importance for detailed garment representation. As shown in Fig.~\ref{fig:ablation_study}(c), removing DiT-GarmentNet compromises fine garment details, like textures and logo placement, indicating its role in preserving intricate design elements. In Fig.~\ref{fig:ablation_study}(d), without Text Preservation Loss, text appears slightly distorted, emphasizing this loss function's role in maintaining text fidelity. As shown in Fig.~\ref{fig:ablation_study}(e), the full model achieves optimal performance by incorporating all components, accurately capturing both structural and stylistic details. Quantitative evaluation in Table~\ref{tab:ablation_study_combined} further supports these observations.

\begin{figure}[t]
    \centering
    \includegraphics[width=1\linewidth]{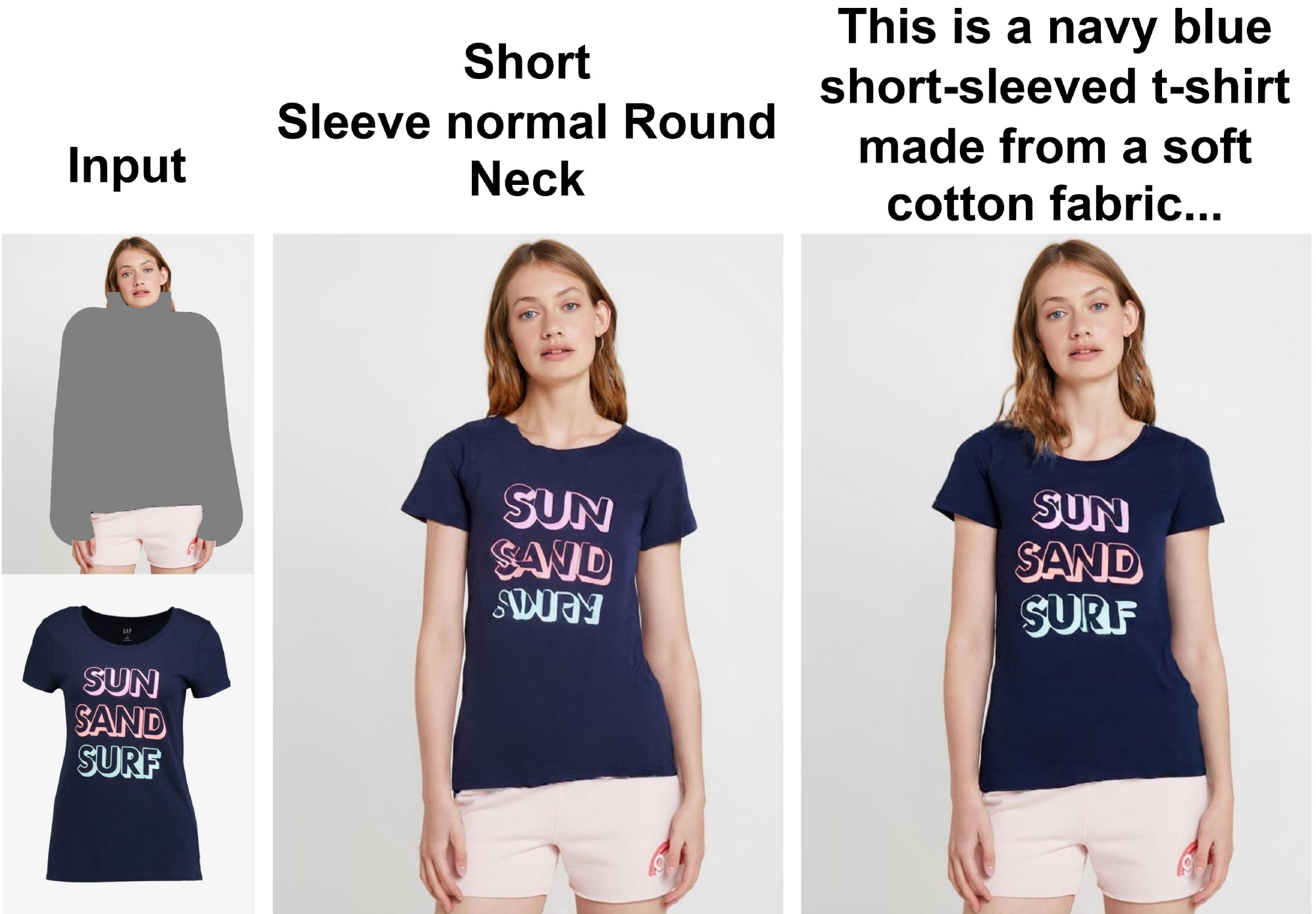}
    \caption{Detailed GPT-generated captions enhance the accuracy of garment-specific details, including color, text, and texture.}
    \label{fig:GPT_caption}
\vspace{-0.2cm}
\end{figure}

\noindent\textbf{Effect of using GPT-generated captions.} We conducted an ablation study to evaluate the effect of detailed GPT-generated captions on TED-VITON’s performance. As shown in Fig.~\ref{fig:GPT_caption}, utilizing GPT-generated captions significantly improves the model’s ability to render multi-line text and garment details like color accuracy and texture. Using a brief description, the model accurately renders the first line, ``SUN'', but fails to capture ``SAND'' and ``SURF'' due to insufficient contextual guidance. Using a detailed caption provides the necessary guidance for accurately rendering all lines and maintaining consistent color and texture. Such quantitative improvements across all metrics are demonstrated in Table~\ref{tab:GPT_caption}.

\begin{table}[t]
    \centering
    
    \vspace{-0.2cm}
    \resizebox{1\linewidth}{!}{
        \begin{tabular}{cccccc}
        \toprule
        \textbf{Detailed Captions} & \textbf{LPIPS↓} & \textbf{SSIM↑} & \textbf{CLIP-I↑} & \textbf{FID↓ UN} & \textbf{KID↓ UN} \\
        \hline
        \ding{55} & 0.113 & 0.872 & 0.829 & 9.881 & 1.706 \\
        \checkmark & \textbf{0.095} & { \textbf{0.881}} & { \textbf{0.878}} & \textbf{8.848} & \textbf{0.858} \\
        \bottomrule
        \end{tabular}
    }
    \caption{Quantitative comparison of models on VITON-HD with and without GPT-generated captions. ``UN'' indicated the unpaired setting.  KID score is multiplied by 100.}
    \label{tab:GPT_caption}
\vspace{-0.4cm}
\end{table}

\section{Conclusion}

We presented TED-VITON, a novel VTO framework built on the DiT architecture to tackle critical challenges in garment detail fidelity and text clarity. By incorporating a GS-Adapter and a Text Preservation Loss, TED-VITON significantly improves garment-specific feature representation and ensures distortion-free rendering of logos and text. Additionally, a constraint mechanism for LLM-generated prompts enhances training inputs, leading to superior performance. Comprehensive evaluations on the VITON-HD \cite{choi_viton-hd_2021} and DressCode \cite{morelli_dress_2022} datasets showcase state-of-the-art results in visual quality, garment alignment, and text fidelity, establishing TED-VITON as a scalable and high-quality solution for next-generation VTO applications. 

{
    \small
    \bibliographystyle{ieeenat_fullname}
    \bibliography{main}
}

\end{document}